\documentclass[letterpaper]{article} 
\usepackage{aaai23}  
\usepackage{times}  
\usepackage{helvet}  
\usepackage{courier}  
\usepackage[hyphens]{url}  
\usepackage{graphicx} 
\usepackage{natbib}  
\usepackage{caption} 
\usepackage{algorithm}
\usepackage{algorithmic}
\usepackage{amsmath}
\usepackage{amsfonts}
\usepackage{multirow}

\usepackage{newfloat}
\usepackage{listings}
\DeclareCaptionStyle{ruled}{labelfont=normalfont,labelsep=colon,strut=off} 
\floatstyle{ruled}
\newfloat{listing}{tb}{lst}{}
\floatname{listing}{Listing}
\title{Towards a Holistic Understanding of Mathematical Questions \\ with Contrastive Pre-training}
\author{
    Yuting Ning,\textsuperscript{\rm 1,\rm 2}
    Zhenya Huang,\textsuperscript{\rm 1,\rm2}
    Xin Lin,\textsuperscript{\rm 1,\rm2}
    Enhong Chen,\textsuperscript{\rm 1,\rm2}\thanks{Corresponding Author.}\\
    Shiwei Tong,\textsuperscript{\rm 1,\rm2}
    Zheng Gong,\textsuperscript{\rm 1,\rm2}
    Shijin Wang\textsuperscript{\rm 2,\rm3}
}
\affiliations{
    \textsuperscript{\rm 1} Anhui Province Key Laboratory of Big Data Analysis and Application,\\
    School of Computer Science and Technology, University of Science and Technology of China\\
    \textsuperscript{\rm 2} State Key Laboratory of Cognitive Intelligence\\
    \textsuperscript{\rm 3} iFLYTEK AI Research (Central China), iFLYTEK Co., Ltd. \\
    \{ningyt, linx, tongsw, gz70229\}@mail.ustc.edu.cn, \{huangzhy, cheneh\}@ustc.edu.cn, sjwang3@iflytek.com
}

\usepackage{bibentry}

\begin{document}

\maketitle

\begin{abstract}
Understanding mathematical questions effectively is a crucial task, which can benefit many applications, such as difficulty estimation. Researchers have drawn much attention to designing pre-training models for question representations due to the scarcity of human annotations (e.g., labeling difficulty). However, unlike general free-format texts (e.g., user comments), mathematical questions are generally designed with explicit purposes and mathematical logic, and usually consist of more complex content, such as formulas, and related mathematical knowledge (e.g., \textit{Function}). Therefore, the problem of holistically representing mathematical questions remains underexplored. To this end, in this paper, we propose a novel contrastive pre-training approach for mathematical question representations, namely QuesCo, which attempts to bring questions with more similar purposes closer. Specifically, we first design two-level question augmentations, including content-level and structure-level, which generate literally diverse question pairs with similar purposes. Then, to fully exploit hierarchical information of knowledge concepts, we propose a knowledge hierarchy-aware rank strategy (KHAR), which ranks the similarities between questions in a fine-grained manner. Next, we adopt a ranking contrastive learning task to optimize our model based on the augmented and ranked questions. We conduct extensive experiments on two real-world mathematical datasets. The experimental results demonstrate the effectiveness of our model.
\end{abstract}

\section{Introduction}

Online learning systems, such as Coursera\footnote{https://www.coursera.org/} and edX\footnote{https://www.edx.org/}, have attracted much attention worldwide in recent years \cite{anderson2008theory, lockee2021online}.
These online learning systems have collected massive educational questions, and provide personalized applications based on the understanding of these questions \cite{benedetto2020introducing}.
For example, we should understand the difficulties of questions to recommend suitable questions for learners with different abilities \cite{benedetto2020r2de}.
Especially, among all educational questions, mathematical questions are more difficult to understand with special components (e.g., formulas) and complex mathematical logic, which require more domain knowledge for comprehensive understanding.
Therefore, mathematical question understanding is a crucial but challenging issue in intelligent education field, which has attracted widespread attention \cite{Davies2021AdvancingMB, zhao2022jiuzhang}.

\begin{figure*}[t]
\centering
\includegraphics[width=0.9\textwidth]{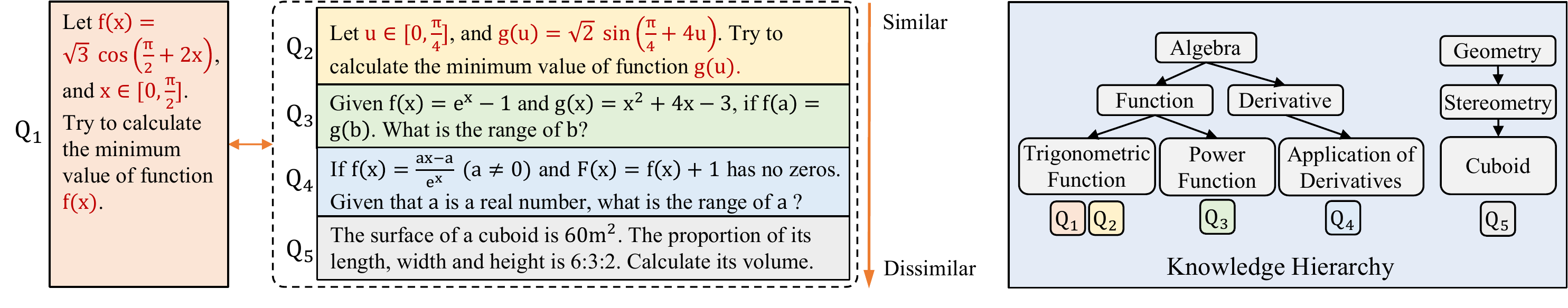} 
\caption{Examples of mathematical questions and knowledge hierarchy. For $Q_1$, the similarity decreases from $Q_2$ to $Q_5$.}
\label{example}
\end{figure*}

In the literature, there are many efforts for mathematical question understanding, including task-specific \cite{ughade2019survey} and pre-training methods \cite{yin2019quesnet}. Task-specific methods focus on one specific aspect of questions (e.g., difficulty), which require massive expertise for human annotations and suffer from the label sparsity \cite{huang2021stan}.
Therefore, pre-training has been utilized for this task and has shown strong effectiveness recently.
Pre-training methods \cite{yin2019quesnet, zhao2022jiuzhang} aim to learn comprehensive question representations on large-scale unlabeled data and benefit various applications (e.g., difficulty estimation).
However, existing methods, such as MathBERT~\cite{peng2021mathbert}, mainly focus on the details of question content, but lack the consideration of holistic meanings from a mathematical perspective, which is more important. As shown in Figure~\ref{example}, although $Q_1$ and $Q_2$ are literally different, they are typically considered mathematically similar, as they have similar holistic purposes with the same mathematical knowledge and similar solutions. Therefore, we hope to design special pre-training methods for understanding holistic meanings of mathematical questions. \looseness=-1

However, there are many challenges along this line.
First, compared with general texts (e.g., user reviews) and other educational questions (e.g., literature questions), mathematical questions are more complex with special components (e.g., formulas), and require more mathematical knowledge and logic to understand.
For example, in Figure~\ref{example}, the formula $f(x)=e^x-1$ implies that the question examines the mathematical knowledge \textit{Power Function}, and greatly affects the solution. Besides, mathematical terms, which are often treated as synonyms generally, may have quite different mathematical properties (e.g., \textit{circle} and \textit{ellipse}). Therefore, pre-training methods for general texts could not be directly applied to mathematical questions.
Second, the holistic mathematical purposes of questions are more important than literal details. For example, in Figure~\ref{example}, although the contents of $Q_1$ and $Q_2$ are quite different (e.g., different variables $u$ and $x$), they are actually mathematically similar since they both target the range of complicated trigonometric functions. However, most existing pre-training methods for mathematical questions, such as MathBERT~\cite{peng2021mathbert}, tend to capture details of formulas but fail on holistic purposes. How to instruct the model to identify holistic purposes of mathematical questions is a challenging issue.
Third, the related knowledge concepts play an important role in understanding mathematical questions, since they reflect the purposes and mathematical domain of questions. The knowledge concepts have complex structures, which contain rich information for question understanding. For example, in Figure~\ref{example}, although $Q_3$ and $Q_4$ both have different concrete knowledge concepts with $Q_1$, $Q_3$ is more similar to $Q_1$ since they share a more fine-grained knowledge concept $Function$. However, the rich information in knowledge concepts is ignored by most existing methods, such as DisenQNet~\cite{huang2021disenqnet}, as they only exploit knowledge concepts as independent ones. How to exploit such information remains open.

To this end, in this paper, we propose a novel contrastive pre-training method for holistically understanding mathematical questions, namely QuesCo.
We specially construct contrastive samples to capture the latent purposes of mathematical questions from a holistic perspective.
First, considering the complex domain-specific features of mathematical questions, we design two-level question augmentations for the content and structure of questions respectively, and further augment the two key components at the content level, including the text and mathematical formulas. In this way, we can generate literally diverse question pairs with similar latent purposes.
Then, to fully exploit hierarchical information of knowledge concepts, we design a novel knowledge hierarchy-aware rank strategy (KHAR), which ranks the similarities between mathematical questions in a fine-grained manner.
After that, we optimize the question representations with ranking contrastive loss based on the augmented and ranked samples.
Finally, we perform extensive experiments on two real-world datasets, which demonstrates that our proposed pre-training method is effective and the learned question representations can benefit various downstream tasks.\looseness=-1

\section{Related Work}
In this section, we review the related works as follows.

\noindent\textbf{Mathematical Question Understanding.}
Mathematical question understanding is a fundamental task in intelligent education, supporting various scenarios, such as question search~\cite{pelanek2019measuring} and personalized recommendation~\cite{liu2019ekt}.
Existing methods could be roughly divided into two categories: task-specific and task-agnostic methods.
Task-specific methods focus on specific aspects of questions.
Earlier works designed rules or grammars \cite{shi2015automatically}, while scholars have tried to utilize supervised deep learning methods recently \cite{lin2021hms, liu2018finding}.
For instance, Huang et al.~\shortcite{huang2020neural} utilized enhanced formula structures to solve mathematical questions.
However, these works require massive expertise for annotations and suffer from the label sparsity \cite{huang2021stan}.
Therefore, many advances try to learn comprehensive representations for mathematical questions with pre-trained language models, which make use of large-scale unlabeled data to benefit different downstream tasks. 
Existing pre-training methods mainly rely on the reconstruction of details. For example, QuesNet~\cite{yin2019quesnet}, MathBERT~\cite{peng2021mathbert} and JiuZhang~\cite{zhao2022jiuzhang} adopt the variants of masked language model.
DisenQNet~\cite{huang2021disenqnet} pre-trained disentangled representations to capture the concepts and individual meanings of questions, while it ignored the complex structure of knowledge concepts.

\noindent\textbf{Pre-trained Language Model.} In recent years, the emergence of pre-trained language models (PLMs) has brought natural language processing (NLP) to a new era~\cite{qiu2020pre}, which has resulted in impressive performances across many NLP tasks~\cite{wu2021context, weng2020acquiring}.
BERT family ~\cite{devlin2019bert, liu2019roberta, lan2019albert} is the most notable one of PLMs. Generally, they utilize Transformer~\cite{vaswani2017attention} together with  pre-training objectives, such as the masked language model (MLM). Although great success has been achieved, they are hard to be directly applied to understand questions, as they cannot capture the domain-specific features of questions, such as formulas and related knowledge concepts.

\begin{figure*}[t]
\centering
\includegraphics[width=0.9\textwidth]{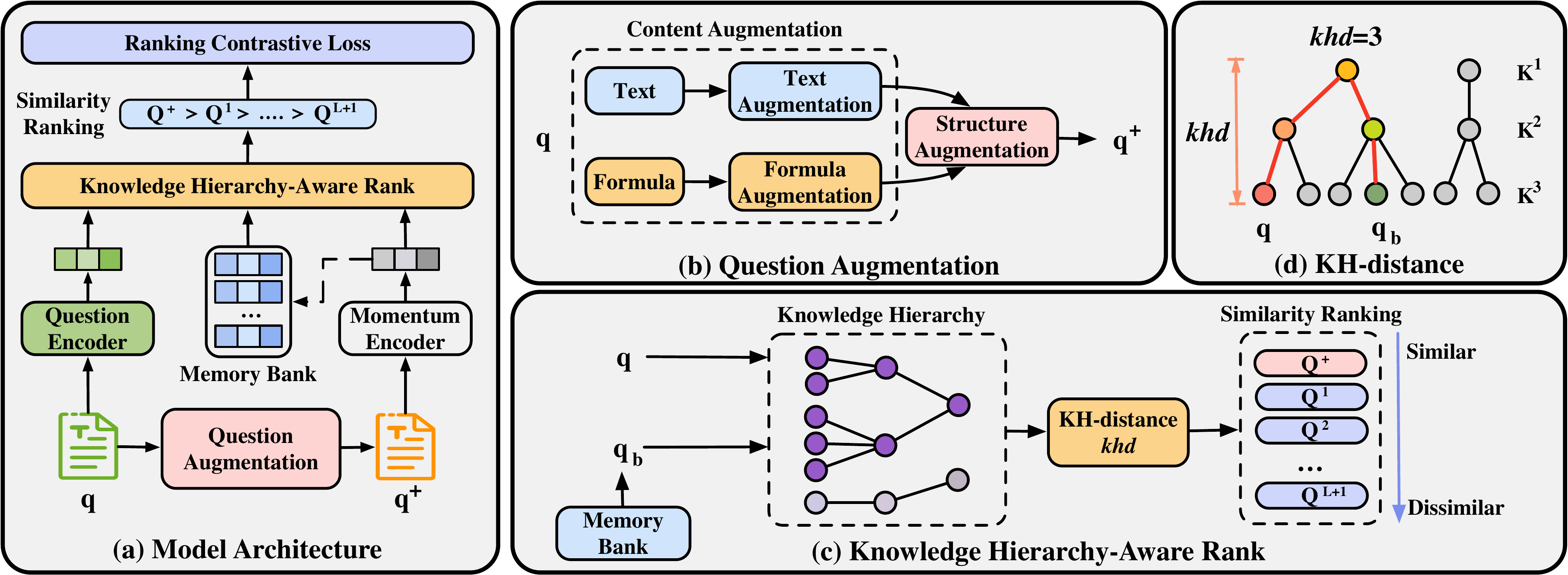} 
\caption{The proposed QuesCo framework. (a) Overall architecture. (b) Question augmentation. (c) Knowledge hierarchy-aware rank. (d) KH-distance calculation.}
\label{framework}
\end{figure*}

\noindent\textbf{Contrastive Learning.}
Contrastive learning 
aims to learn effective representations by identifying whether samples are holistically similar~\cite{chen2020simple, he2020momentum}.
There are two key problems.
The first one is the construction of similar (positive) pairs and dissimilar (negative) pairs. Unsupervised methods use data augmentations to generate different views of samples, which are treated as positives, while other samples are negatives \cite{yan2021consert}.
These methods benefits from large batch sizes \cite{chen2020simple} or memory bank \cite{he2020momentum, liu2022learning} with more negatives.
In supervised settings, samples from the same class are regarded as positives~\cite{khosla2020supervised}.
The other key problem is the contrastive loss. InfoNCE~\cite{oord2018representation}, the most widely used loss function, maximizes the similarity of positive pairs and minimizes the similarity of negative pairs. Besides, \citeauthor{hoffmann2022ranking}\shortcite{hoffmann2022ranking} proposed a ranking contrastive loss RINCE, which replaced the binary definition of positives and negatives with a ranked similarity.
Contrastive learning has been applied to many NLP tasks and shown strong effectiveness~\cite{giorgi2021declutr, xu2022sequence, fu2021lrc, pan2022improved, wang2022incorporating}.

Our work provides a holistic representation for mathematical questions, which differs from previous methods as follows. First, we consider complex math features (e.g., formulas), which are neglected in general pre-training methods. Second, we try to learn holistic purposes of questions, while existing question pre-training tasks focus more on details. Third, we try to exploit rich information in hierarchical knowledge concepts, which is ignored by existing methods.

\section{QuesCo Framework}

In this section, we introduce our proposed pre-training method QuesCo for mathematical questions. We first formally define the mathematical questions and present the question representation problem. Then we introduce the QuesCo architecture and pre-training approach.

\subsection{Problem Definition}

In this subsection, we formally introduce the mathematical questions and the question representation problem.

\subsubsection{Mathematical Questions.}
Mathematical questions consist of content and related knowledge concepts $ \boldsymbol{q} = \left( \boldsymbol{x}, \boldsymbol{k}\right)$. Question content is a sequence of $T$ tokens, denoted as $\boldsymbol{x}= \{x_1, x_2, \cdots, x_T\}$, where each token $x_i$ is either a text  word or a formula symbol.
Related knowledge concepts are selected from a $L$-level knowledge hierarchy $KH=\{\mathcal{K}, \mathcal{E} \}$.
As shown in Figure~\ref{example}, $KH$ is a tree structure, where the vertexes $\mathcal{K}$ are the knowledge concepts, and the edges $\mathcal{E}$ represent the relationship between knowledge concepts.
We denote the set of knowledge concepts at the $\ell$-level as $\boldsymbol{K^\ell} = \{k^\ell_1, k^\ell_2, \cdots, k^\ell_{n_\ell}\} \subseteq \mathcal{K}$. The smaller $\ell$ indicates a higher knowledge level, which represents a more coarse-grained knowledge (e.g. \textit{Function}).
In this tree structure, $(k^\ell_i, k^{\ell+1}_j) \in \mathcal{E}$ means that the child $k^{\ell+1}_j$ is a more fine-grained knowledge of $k^{\ell}_i$, e.g. \textit{Power Function} represents a specific part of \textit{Function} in Figure~\ref{example}.
Therefore, the related knowledge concepts of question $\boldsymbol{q}$ are denoted as $\boldsymbol{k} = \{k^1, k^2, \cdots, k^L\} $ , where $k^i\in \boldsymbol{K^i}$ and $k^i$ is the ancestor of $k^j (j > i)$. \looseness=-1

\subsubsection{Question Representation Problem.}

Given a mathematical question $ \boldsymbol{q} = \left( \boldsymbol{x}, \boldsymbol{k}\right)$, our goal is to represent $\boldsymbol{q}$ with a $d$-dimensional vector $\boldsymbol{v} \in \mathbb{R}^d$, which can be transferred to several downstream tasks and benefit their performances, such as difficulty estimation. We expect the learned representations to capture latent purposes of questions from a holistic perspective, and contain the rich information in question content (e.g., formulas) and related knowledge concepts.

\subsection{Model Overview}
As discussed above, the holistic purposes are more important than literal details in mathematical question understanding.
However, existing question pre-training methods, which mainly rely on some variants of the masked language model, are more efficient for token-level details but fail on holistic purposes.
Comparatively, contrastive learning has shown strong effectiveness in learning holistic representations \cite{yan2021consert, gao2021simcse}.

Based on contrastive learning, we propose a novel pre-training approach for mathematical questions, namely QuesCo.
Figure~\ref{framework}(a) depicts the framework of QuesCo.
We aim to learn comprehensive question representations by pulling questions with more similar purposes closer than those with less similar purposes.
Thus, the most important issue is the construction of contrastive pairs that are consistent with the mathematically holistic similarity.
Specifically, for a given question (i.e., anchor question), we propose question augmentation strategies (Figure~\ref{framework}(b)) to generate literally diverse questions with similar holistic purposes, which perform two-level augmentations for content (i.e., the text and formulas) and structure of questions respectively. Furthermore, based on the rich information in knowledge concepts, which imply holistic purposes of questions, we design the Knowledge Hierarchy-Aware Rank (KHAR) module (Figure~\ref{framework}(c)) to fully exploit fine-grained similarity ranking of other questions and the anchor question.
Then, based on the similarity ranking in KHAR module, we use a ranking contrastive loss to learn question representations.
Besides, following He et al.~\shortcite{he2020momentum}, we introduce a memory bank with a momentum encoder, which allows us to store a large set of questions and their knowledge concepts. \looseness=-1

In the following subsections, we will introduce two key components and pre-training method of QuesCo in detail.

\subsection{Question Augmentation}

In contrastive pre-training, to learn latent purposes of mathematical questions, the similar pairs should share similar holistic purposes but have diverse literal details, which are often generated by data augmentation.
However, as shown in Figure~\ref{example}, mathematical question content is complex with various components (i.e., plain text and formulas) and mathematical knowledge.
Especially, the formulas, which consist of math symbols, contain much mathematical knowledge and logic.
Therefore, existing data augmentations can not be directly applied, which requires us to design augmentations for mathematical question content.
Moreover, question structure also has unique logic (e.g., parallelism of conditional clauses), which helps to understand questions.
Thus, as shown in Figure~\ref{framework}(b), we propose two-level question augmentations, i.e. \textit{content} and \textit{structure}.
We introduce our proposed two-level data augmentations as follows. 

As for question content, formulas and texts require different domain logic (e.g., equivalent transformation of variables). Therefore, at the content level, we further augment \textit{text} and \textit{formula} respectively.

\subsubsection{Text Augmentation.}
There are many augmentations designed for texts in general NLP tasks.
Here we adopt two commonly used strategies for plain texts which would not change the meaning greatly, i.e., random swap and random deletion \cite{wei2019eda}. The former one randomly chooses two words and swaps their positions, and the latter randomly removes a word.
Note that not all strategies for general texts can be directly applied, such as strategies that leverage the synonym dictionary.
Many mathematical terms, which are commonly treated as synonyms in general texts, may have different properties in questions (e.g. \textit{circle} and \textit{ellipse}). Treating them as synonyms for augmentation will cause damage to mathematical question understanding.

\subsubsection{Formula Augmentation.}
As symbols in mathematical formulas contain rich mathematical logic, we need to design special augmentation strategies for formulas. There are three main types of mathematical symbols in formulas, i.e., variables, operators and numbers. Therefore, we design the following strategies for these three components.

$\bullet$ \textbf{Variable Renaming.} The identifiers of variables do not affect the meanings of formulas. We randomly rename one variable in the question with another unused one. For example, $f(x)=\cos{(\frac{\pi}{2}+2x)}$ and $f(u)=\cos{(\frac{\pi}{2}+2u)}$ are actually the same mathematically.

$\bullet$ \textbf{Variable Scaling.} We randomly select one variable and perform arithmetic operations on it, e.g. expand $x$ by a factor of $2$ in $f(x)=\cos{(\frac{\pi}{2}+2x)}$ and get a new formula $f(x)=\cos{(\frac{\pi}{2}+4x)}$ . The answer may differ because of this transformation, while the solution remains quite similar.

$\bullet$ \textbf{Operator Synonym Replacement.} We notice that many operators have similar mathematical properties, which can be treated as synonyms, such as $\sin$ and $\cos$. By replacing a randomly selected operator with its mathematical synonym, we can generate a similar question whose required mathematical knowledge remains essentially the same.

$\bullet$ \textbf{Number Replacement.} The specific values of numbers merely affect the methods to solve questions. Thus, we recognize the numbers in formulas and randomly replace one.

\subsubsection{Structure Augmentation.}
In addition to content, question structure also has unique properties. Therefore, as shown in Figure~\ref{framework}(b), after content-level augmentations, we further perform augmentations for question structure. Here we propose two strategies for question structure.

$\bullet$ \textbf{Clause Shuffling.} Mathematical questions usually consist of several conditional clauses, which are often in a parallel relationship. The meaning and solutions of questions are insensitive to the order of these parallel clauses. Therefore we can shuffle the conditional clauses for augmentation.

$\bullet$ \textbf{Useless Clause Insertion.} Inserting irrelevant conditions into a question will affect its semantics, while its purpose and solution remain unchanged. However, it is hard to judge whether a random insertion is irrelevant. Here we randomly select one existing conditional clause for repetition and insert it into a random position between current clauses.

Each strategy will be applied with the probability $p$ to produce a diverse set of augmented questions. Note that though we focus on mathematical questions here, these strategies could be generalized to all educational questions.
For example, we could also perform text, formula and structure augmentations for physical questions and design more strategies based on other characteristics, such as physical theories.

\subsection{Knowledge Hierarchy-Aware Rank}

As mentioned above, knowledge concepts are important to understand mathematical questions, as they reflect the mathematical domain and holistic purposes. For instance, as shown in Figure~\ref{example}, $Q_1$ and $Q_2$ are mathematically quite similar as they share the same concrete knowledge \textit{Trigonometric Function}, while $Q_1$ and $Q_5$ are dissimilar with different knowledge concepts. Therefore, we could exploit such information in knowledge concepts to construct contrastive pairs.

Knowledge concepts are not independent and contain rich information in their complex structure, thus mathematical question pairs cannot be simply divided into positives and negatives based on them just as many traditional contrastive learning methods do.
For example, as shown in Figure~\ref{example}, although $Q_3$ and $Q_4$ both have different concrete knowledge concepts from $Q_1$, $Q_3$ is more similar to $Q_1$. This is because $Q_3$ and $Q_1$ share a more fine-grained knowledge \textit{Function}, while the shared knowledge \textit{Algebra} of $Q_4$ and $Q_1$ is more coarse-grained.
Motivated by this observation, we exploit fine-grained similarities between questions based on the relationship of mathematical knowledge concepts.

As shown in Figure~\ref{framework}(c), we propose the Knowledge Hierarchy-Aware Rank (KHAR) module to utilize the hierarchical relationship of mathematical knowledge concepts. We divide mathematical questions into different similarity ranks according to their distance with the anchor question in the knowledge hierarchy. Formally, we define the distance between question $q_i$ and $q_j$ in an $L$-level knowledge hierarchy as the KH-distance
\begin{equation}
\small
    \textit{khd}(q_i, q_j) =
    \left\{
    \begin{array}{ll}
    L - u + 1 & \text{if $\exists u \in \{1, \cdots, L\}$,}\\
    &\text{ $k^u_i = k^u_j$ and $k^{u+1}_i \neq k^{u+1}_j$} \\
    L + 1 & \text{if $\forall \ell \in \{1, \cdots, L\}, k^\ell_i \neq k^\ell_j$}
    \end{array}
    \right.
    ,
\label{khd}
\end{equation}
where $k^u_i\in \boldsymbol{k_i}$ and $k^u_j\in \boldsymbol{k_j}$ represent the most fine-grained shared knowledge concept of $q_i$ and $q_j$ if any.
As shown in Figure~\ref{framework}(d), KH-distance indicates the distance from the bottom to the shared knowledge concept of a question pair in the knowledge hierarchy, which represents their similarity.
For instance, in Figure~\ref{framework}(d), in the 3-level knowledge hierarchy, the most fine-grained shared knowledge concept of $Q$ and $Q_b$ lies on the 1st level, i.e., $u=1$. According to Eq.(\ref{khd}), their KH-distance is 3.
A smaller distance indicates that they require more fine-grained shared knowledge and are more similar.
Specially, $\textit{khd}(q_i, q_j) = L+1$ indicates that they are completely dissimilar as they have totally different knowledge concepts.

Given an anchor question $q$, we can assign each question $q_i$ in memory bank into one of $L+1$ ranks according to the $\textit{khd}(q, q_i) \in \{1, \cdots, L+1\}$. 
Besides, as our proposed question augmentations do not change the holistic purposes of questions, the augmented question $q^+$ is the most similar to the anchor question, which can be specially denoted as $\textit{khd}(q, q^+)=0$. Therefore, we can obtain the similarity ranking for a given question $q$,
\begin{equation}
\small
\label{rank}
    h(q, q^0) > h(q, q^1) > \cdots > h(q,q^{L+1}), \forall q^u \in Q^u
    ,
\end{equation}
where $h$ indicates the similarity function.
Here $Q^u$ denotes the question set of questions whose distance with the anchor question is $u$ and $u\in\{0, \cdots, L+1\}$:
\begin{equation}
\small
\label{ques-set}
    Q^u = 
    \left\{
    \begin{array}{cl}
        \{q^+\},& u=0 \\
        \{p | \textit{khd}(q, p) = u\}, & u\in \{1, \cdots, L+1\}
    \end{array}
    \right.
    .
\end{equation}
According to Eq.(\ref{rank}) and Eq.(\ref{ques-set}), we fully exploit the hierarchical information of knowledge concepts, and obtain a ranking to capture the fine-grained similarities of contrastive samples for a given anchor question.

\subsection{Pre-training}
With question augmentations and KHAR module, for an anchor question $q$, we can obtain contrastive pairs with a fine-grained similarity ranking in Eq.(\ref{rank}).
Based on that, we adopt the Ranking Info Noise Contrastive Estimation loss (RINCE)~\cite{hoffmann2022ranking} as our contrastive objective, which can exploit more information. More precisely, the loss function reads $L_{rank} = \sum^L_0 \ell_i$, where
\begin{equation}
\small
    \ell_i = -\log \frac{\sum_{p\in Q^i} \exp{(\frac{h(q, p)}{\tau_i}})}
    {\sum_{p\in \bigcup_{j\geq i}Q^j} \exp{(\frac{h(q, p)}{\tau_j}})}
    .
\end{equation}
Here $Q^i$ is the $i$-th question set defined in Eq.(\ref{ques-set}).
We use dot product similarity on question representation $\boldsymbol{v}$ from the question encoder as $h(\cdot)$.
Following \citep{chen2020simple}, the question encoder consists of a BERT encoder and a projector. We use the mean of token embeddings at the last layer of BERT encoder, and map them to the space where contrastive loss is applied with the projector.
Besides, $\tau_i$ is a hyper-parameter which controls the temperature and $\tau_i < \tau_{i+1}$.

RINCE is a variant of the widely-used loss, i.e., infoNCE. Initially, we treat the first question set $Q^0$ (i.e., the augmented question) as positives while questions with lower ranks (i.e., $\bigcup_{j>0}Q^j$) are negatives, and calculate infoNCE loss as $\ell_0$. Then we drop $Q^0$ and move to $Q^1$.
we repeat this procedure until only $Q^{L+1}$ left, where questions have no shared knowledge concept with the anchor question.

In this way, QuesCo enforces gradually decreasing similarity with increasing rank of samples. We can exploit more information based on the similarity ranking for learning effective mathematical question representations.

\section{Experiment}

In this section, we conduct experiments on three commonly used tasks with two real-world education datasets, to verify the effectiveness of our proposed approach.

\subsection{Experimental Setup}

\subsubsection{Datasets.}

\begin{table}[t]
    \small
    \centering
    \setlength{\tabcolsep}{0.9 mm}{
    \begin{tabular}{c|c|c}
    \hline\hline
    Statistics & SYSTEM1 & SYSTEM2 \\
    \hline
    \# Questions & 123,200 & 25,287 \\
        Avg. question length & 80.08 & 130.91 \\
        Avg. formula length per question & 48.02 & 84.48 \\
        \hline
    \# Hierarchical levels & 3 & 3 \\
        \# Knowledge in level-1 & 21 & 21 \\
        \# Knowledge in level-2 & 81 & 54 \\
        \# Knowledge in level-3 & 361 & 175 \\
        \hline
        \# Questions with difficulty label & 7,056 & / \\
        \# Questions with similarity label & / & 6,873 \\
        Label sparsity & 5.72\% & 27.18\% \\
    \hline\hline
    \end{tabular}}
    \caption{The statistics of the datasets.}\label{tab:data}
\end{table}

\begin{figure}[t]
\centering
\includegraphics[width=0.8\columnwidth]{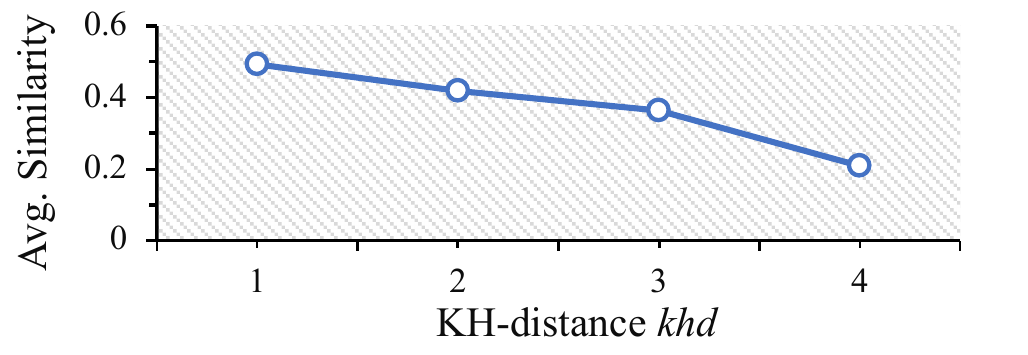} 
\caption{The relationship between $khd$ and labeled similarity between questions in SYSTEM2.}
\label{similarity}
\end{figure}

\begin{table*}[t]
    \centering
    \small
    \setlength{\tabcolsep}{1.05 mm}{
    \begin{tabular}{c|c|c|c|c|c|c|c|c|c|c|c|c|c|c}
    \hline\hline
    Tasks & \multicolumn{2}{c|}{Similarity Prediction} &
    \multicolumn{8}{c|}{Concept Prediction} & \multicolumn{4}{c}{Difficulty Estimation} \\
    \hline
    Datasets & \multicolumn{2}{c|}{SYSTEM2} &
    \multicolumn{4}{c|}{SYSTEM1} & \multicolumn{4}{c|}{SYSTEM2} & \multicolumn{4}{c}{SYSTEM1} \\
    \hline
    \multirow{2}{*}{Metrics} & \multirow{2}{*}{Pearson} & \multirow{2}{*}{Spearman} &
    \multicolumn{2}{c|}{level-1} & \multicolumn{2}{c|}{level-2} & \multicolumn{2}{c|}{level-1} & \multicolumn{2}{c|}{level-2} &
    \multirow{2}{*}{MAE} & \multirow{2}{*}{RMSE} & \multirow{2}{*}{PCC} & \multirow{2}{*}{DOA} \\
    \cline{4-11}
     & & & ACC & F1 & ACC & F1 & ACC & F1 & ACC & F1 & & & \\
     \hline
     BERT & 0.2957 & 0.3655 & 0.7309 & 0.5213 & 0.4472 & 0.1833 & 0.4822 & 0.2374  & 0.2984 & 0.0945 & 0.1987 & 0.2463 & 0.3974 & 0.6318 \\
     DAPT-BERT & 0.4856 & 0.5313 & 0.8032 & 0.6288 & 0.5597 & 0.2727 & 0.6522 & 0.3855 & 0.4960 & 0.1836 & 0.1880 & 0.2313 & 0.5087 & 0.6589 \\
     ConSERT & 0.5060 & 0.4760 & 0.8064 & 0.6655 & 0.5933 & 0.3135 & 0.6987 & 0.4952 & 0.5020 & 0.2076 & 0.1873 & 0.2308 & 0.5115 & 0.6621 \\
     SCL & 0.6901 & 0.7101 & 0.8985 & 0.8011 & 0.7492 & 0.4683 & 0.8083 & 0.6498 & 0.6225 & 0.3071 & 0.1996 & 0.2460 & 0.4002 & 0.6340 \\
     QuesNet & 0.5370 & 0.5549 & 0.7881 & 0.6930 & 0.5693 & 0.3604 & 0.7194 & 0.6213 & 0.5810 & 0.3118 & 0.1865 & 0.2305 & 0.3959 & 0.6539 \\
     DisenQNet & 0.6922 & 0.6955 & 0.8210 & 0.7064 & 0.6404 & 0.4332 & 0.7945 & 0.6805 & 0.2431 & 0.1023 & 0.1970 & 0.2424 & 0.4293 & 0.6338 \\
     QuesCo & \textbf{0.7385} & \textbf{0.7245} & \textbf{0.9176} & \textbf{0.8938} & \textbf{0.7857} & \textbf{0.5550} & \textbf{0.8340} & \textbf{0.7018} & \textbf{0.6719} & \textbf{0.3756} & \textbf{0.1778} & \textbf{0.2219} & \textbf{0.5623} & \textbf{0.6765} \\
     \hline\hline
    \end{tabular}}
    \caption{Performance comparisons on different tasks. All the improvements over baselines are significant at $p<0.01$.}\label{tab:tasks}
\end{table*}

We use two real-world datasets in experiments, namely SYSTEM1 and SYSTEM2. The SYSTEM1 dataset collects mathematical questions from an online learning system Zhixue\footnote{http://www.zhixue.com}, which provides various exercise-based applications for high school students in China. The SYSTEM2 dataset collects high school mathematical questions from exams and textbooks, and is labeled by experts. The knowledge hierarchies of both datasets are defined by educational experts. All the formulas are in \LaTeX{} format and make up a large part of question content. Some important statistics are shown in Table~\ref{tab:data}.

In SYSTEM1, following~\cite{broer2005ensuring}, we calculate the difficulty scores of questions, which refer to the correct rates of students, and get 7,056 questions labeled.
In SYSTEM2, we invite three experts to label similarities of question pairs. When determining the similarities of question pairs, experts are expected to consider the similarity of their related knowledge, required ability and quality, based on which the labeled difficulties could be highly consistent among experts with the intraclass correlation coefficient of 0.797. We use the average of their scores as our similarity labels.
We further analyze similarity labels. As shown in Figure~\ref{similarity}, we can observe that the similarity ranking is consistent with our hypothesis, i.e., a smaller $khd$ indicates that questions are likely to be more similar.

\subsubsection{Evaluation Tasks.}

We use three typical tasks based on the mathematical questions, i.e., similarity prediction, concept prediction and difficulty estimation.

We use the zero-shot similarity prediction task to investigate whether the learned similarities are consistent with reality. We directly use the learned question representations without further training. We report Pearson Correlation and Spearman Correlation between calculated cosine similarity scores of representations and the human-annotated gold scores. As the SYSTEM1 dataset does not have similarity labels, we only conduct this task on the SYSTEM2 dataset.

Concept prediction maps a given question to its corresponding knowledge concept. We adopt the widely-used linear evaluation protocol in this task, which trains a linear classifier on top of frozen question representations.
Since hundreds of knowledge concepts in level-3 are hard to distinguish, we evaluate this classification task only on the first two knowledge levels with accuracy and F1 score as metrics. \looseness=-1

Difficulty estimation is a regression task to estimate the difficulty of a given question. We also use the linear evaluation protocol here. We only conduct this task on the SYSTEM1 dataset where we could obtain difficulty scores. Following \cite{huang2017question}, we adopt MAE, RMSE, PCC (Pearson Correlation Coefficient) and DOA (Degree of Agreement) as our evaluation metrics.

\subsubsection{Comparison Methods.}

We compare our proposed approach with the following baselines.

$\bullet$ \textbf{BERT} \cite{devlin2019bert} is the most popular pre-trained model in NLP tasks.

$\bullet$ \textbf{DAPT-BERT} \cite{gururangan2020don} further pre-trains BERT on our pre-training corpus using the MLM task.

$\bullet$ \textbf{ConSERT} \cite{yan2021consert} is a self-supervised contrastive learning method with various augmentation strategies for sentences. 

$\bullet$ \textbf{SCL} \cite{khosla2020supervised} is a supervised contrastive learning method. Here we utilize the level-3 knowledge concepts to construct positives and negatives.

$\bullet$ \textbf{QuesNet} \cite{yin2019quesnet} is the first pre-trained model for questions, which considers heterogeneous inputs including content text, images, and side information. We use it with only question text.

$\bullet$ \textbf{DisenQNet} \cite{huang2021disenqnet} is an unsupervised model for question representation, disentangling questions into the concept and individual representations.


\subsubsection{QuesCo Setup.\footnote{Code is available at https://github.com/bigdata-ustc/QuesCo.}}

\begin{table*}[t]
    \centering
    \small
    \setlength{\tabcolsep}{1.05 mm}{
    \begin{tabular}{c|c|c|c|c|c|c|c|c|c|c|c|c|c|c}
    \hline\hline
    Tasks & \multicolumn{2}{c|}{Similarity Prediction} &
    \multicolumn{8}{c|}{Concept Prediction} & \multicolumn{4}{c}{Difficulty Estimation} \\
    \hline
    Datasets & \multicolumn{2}{c|}{SYSTEM2} &
    \multicolumn{4}{c|}{SYSTEM1} & \multicolumn{4}{c|}{SYSTEM2} & \multicolumn{4}{c}{SYSTEM1} \\
    \hline
    \multirow{2}{*}{Metrics} & \multirow{2}{*}{Pearson} & \multirow{2}{*}{Spearman} &
    \multicolumn{2}{c|}{level-1} & \multicolumn{2}{c|}{level-2} & \multicolumn{2}{c|}{level-1} & \multicolumn{2}{c|}{level-2} &
    \multirow{2}{*}{MAE} & \multirow{2}{*}{RMSE} & \multirow{2}{*}{PCC} & \multirow{2}{*}{DOA} \\
    \cline{4-11}
     & & & ACC & F1 & ACC & F1 & ACC & F1 & ACC & F1 & & & \\
     \hline
     QuesCo & \textbf{0.7385} & \textbf{0.7245} & \textbf{0.9176} & \textbf{0.8938} & \textbf{0.7857} & \textbf{0.5550} & \textbf{0.8340} & \textbf{0.7018} & \textbf{0.6719} & \textbf{0.3756} & \textbf{0.1778} & \textbf{0.2219} & \textbf{0.5623} & \textbf{0.6765} \\
     w/o AUG & 0.7028 & 0.7213 & 0.9079 & 0.8770 & 0.7305 & 0.4497 & 0.8320 & 0.6972 & 0.6443 & 0.3412 & 0.2007 & 0.2482 & 0.3797 & 0.6204 \\
     w/o KHAR & 0.5481 & 0.5057 & 0.8202 & 0.7160 & 0.6181 & 0.3416 & 0.7332 & 0.5996 & 0.5613 & 0.2746 & 0.1810 & 0.2248 & 0.5475 & 0.6655 \\
     \hline\hline
    \end{tabular}}
    \caption{Ablation experiments on different modules.}\label{tab:ablation}
\end{table*}

Our model is implemented by PyTorch. Here we use BERT-Base-Chinese\footnote{https://huggingface.co/bert-base-chinese} as the base encoder. The projector outputs vectors of size 128. The momentum encoder is updated with a momentum term $\mathrm{m}=0.999$. The size of memory bank is set to 1,600. Each augmentation strategy is applied with the probability of $p=0.3$. We adopt the AdamW optimizer with a learning rate of 0.00005.
The temperature factors $\tau_i$ of different ranks in RINCE loss are set to $\{0.1, 0.1, 0.225, 0.35, 0.6\}$ respectively.
We randomly select 10\% questions for concept prediction in each dataset. Then we mix and shuffle the remaining questions of two datasets for pre-training of all methods. For downstream tasks, we randomly partition labeled questions into training/test sets with 80\%/20\%. All experiments are conducted with one Tesla V100 GPU.

\subsection{Overall Results}

The results of three tasks are shown in Table~\ref{tab:tasks}.
Our proposed QuesCo performs consistently better than baselines on all tasks, which are significant at $p<0.01$. This proves that QuesCo obtains a holistic understanding of mathematical questions, which can be transferred to various educational tasks.
There are more observations for different tasks.
First, when predicting similarities, QuesCo reaches the best compared with all baselines. This demonstrates that the modeling of the complex similarity relationship based on question augmentations and knowledge concepts in QuesCo is effective. Besides, we also achieve better performance than contrastive learning baselines in NLP (i.e., ConSERT and SCL), as these baselines ignore the characteristics of mathematical questions.
Second, in concept prediction, QuesCo also performs better than knowledge-enhanced baselines (i.e., SCL and DisenQNet), especially on the fine-grained level. It indicates the effectiveness of capturing subtle differences when predicting concepts by introducing the knowledge hierarchy.
Third, performances of knowledge-enhanced baselines are hardly satisfactory for estimating the difficulty, while QuesCo still achieves the expected results.
It proves that the domain knowledge incorporated by question augmentations is efficient in capturing the difficulty embedded in diverse questions. \looseness=-1

\subsection{Model Analysis}

\begin{figure}[t]
\centering
\includegraphics[width=0.9\columnwidth]{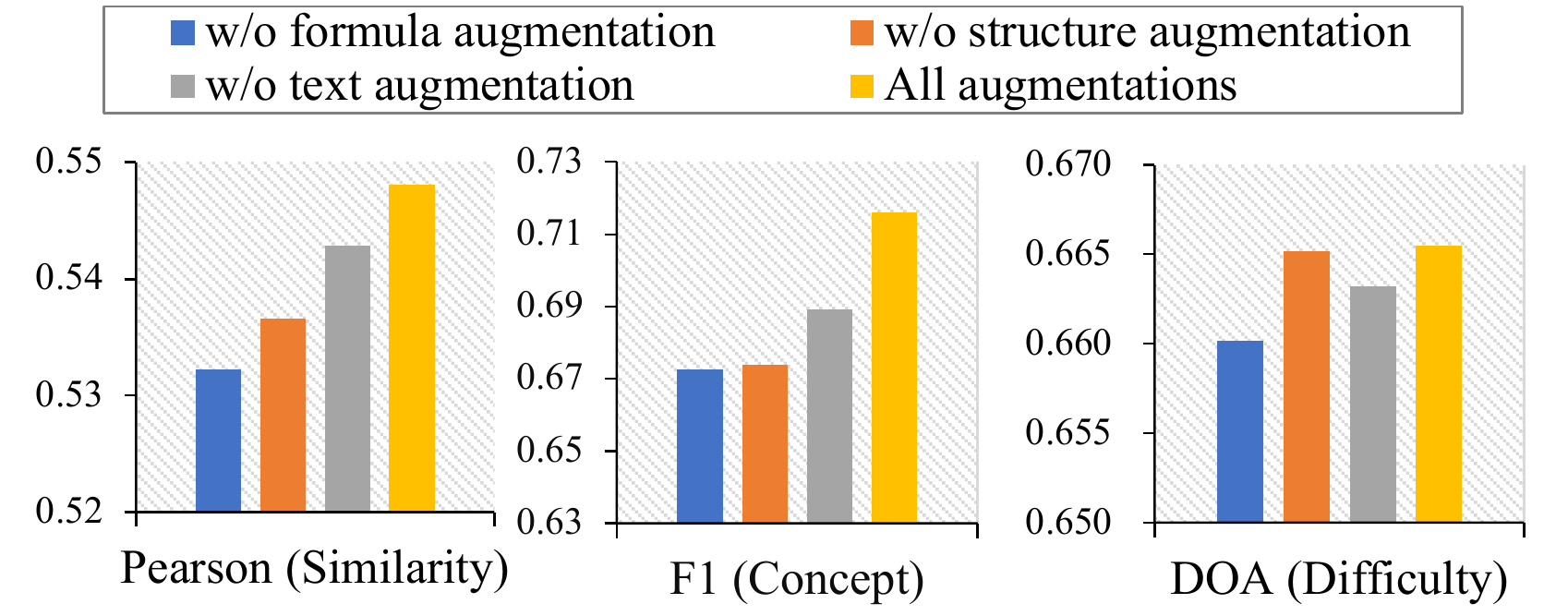} 
\caption{Ablation experiments on different augmentations.}
\label{aug-ablation}
\end{figure}

\subsubsection{Ablation Study.}

To investigate the effect of each component, we conduct ablation studies.
We denote question augmentations as AUG here and the results are shown in Table~\ref{tab:ablation}.
We observe that removing any module will lead to a performance decrease on downstream tasks, which indicates the effectiveness of our designs. Moreover, we notice that the KHAR module is more effective in similarity and concept prediction. It implies that the knowledge hierarchy could help much to understand the similarities and required knowledge of mathematical questions. In contrast, question augmentations are more efficient for capturing the information of difficulty, as we generate questions with similar difficulties but diverse details, which help the model to learn deeper difficulty information behind the content. \looseness=-1

We further study the effect of each augmentation.
To accurately analyze their differences and avoid the interference of other factors, we only use augmented questions as positives in this part. We remove three types of strategies individually. Due to the limitation of space, we only report one metric for each task and the performance on SYSTEM1 for concept prediction in Figure~\ref{aug-ablation}. We observe that performance decreases when any group of augmentation strategies is removed. It demonstrates that all strategies are effective. Among all augmentations, the performance drops the most when removing formula augmentation in all tasks, which implies that it is essential to learn the equivalent transformations of formulas for a deeper understanding of questions.
Besides, as text implies the reading difficulty, the performance of difficulty prediction declines more when text augmentation is removed. In contrast, structure is more important for holistic understanding instead of details, and has more effect on similarity and concept prediction.

\subsubsection{Similarity Ranking Analysis.}

\begin{figure}[t]
\centering
\includegraphics[width=1.0\columnwidth]{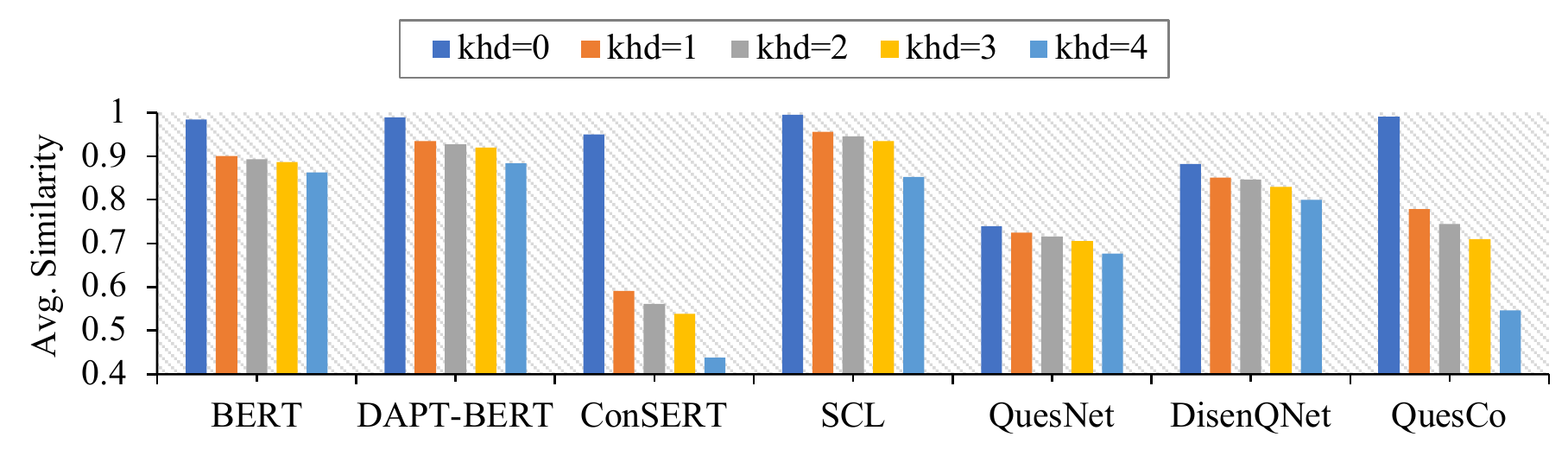} 
\caption{Relationship between $khd$ and similarity.}
\label{sim-rank}
\end{figure}

To analyze whether our model capture fine-grained similarities between questions, we calculate the similarities between questions with different KH-distance which is defined in Eq.(\ref{khd}), and report results in Figure~\ref{sim-rank}. 
Although all the methods can obtain the expected similarity ranking, we can clearly see that QuesCo can better distinguish the fine-grained similarities. The differences of similarities obtained by QuesCo are more obvious across different ranks.


\subsubsection{Visualization.}

\begin{figure}[t]
\centering
\includegraphics[width=1.0\columnwidth]{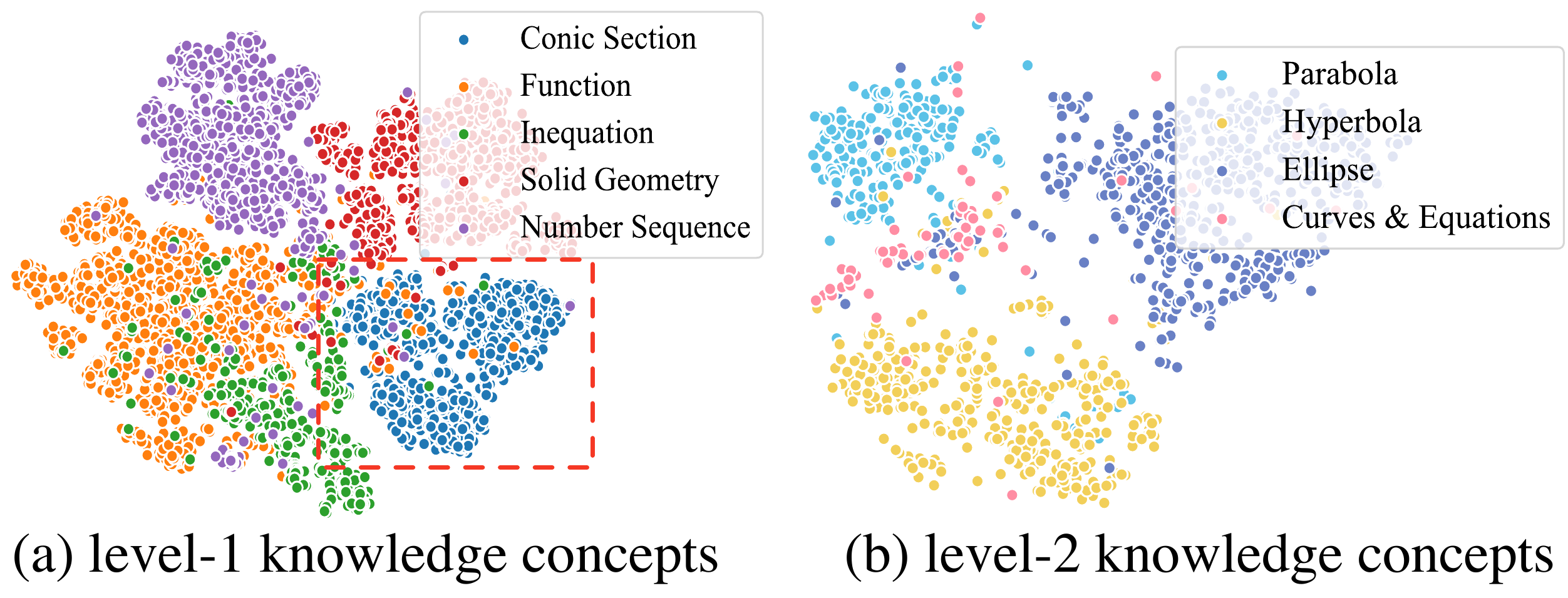} 
\caption{Visualization of questions with knowledge concepts on different levels.}
\label{tsne}
\end{figure}

QuesCo is expected to understand the required knowledge of questions on different levels.
Here we intuitively demonstrate such representation ability.
We choose 5 frequent concepts and their corresponding questions in SYSTEM2. Then we project their question representations into 2D space by t-SNE~\cite{van2008visualizing}.
We first mark questions with level-1 knowledge concepts using different colors in Figure~\ref{tsne}(a). Generally, questions with same concepts are easy to be grouped, demonstrating that QuesCo can capture the required knowledge of questions.
Furthermore, there are also small clusters in each cluster. 
We further mark the level-2 knowledge concepts for questions with \textit{Conic Section} in Figure~\ref{tsne}(b), where questions with the same level-2 concepts are grouped.
This further demonstrates that QuesCo is effective in capturing the knowledge of different levels.

\subsubsection{Case Study.}

\begin{figure}[t]
\centering
\includegraphics[width=1.0\columnwidth]{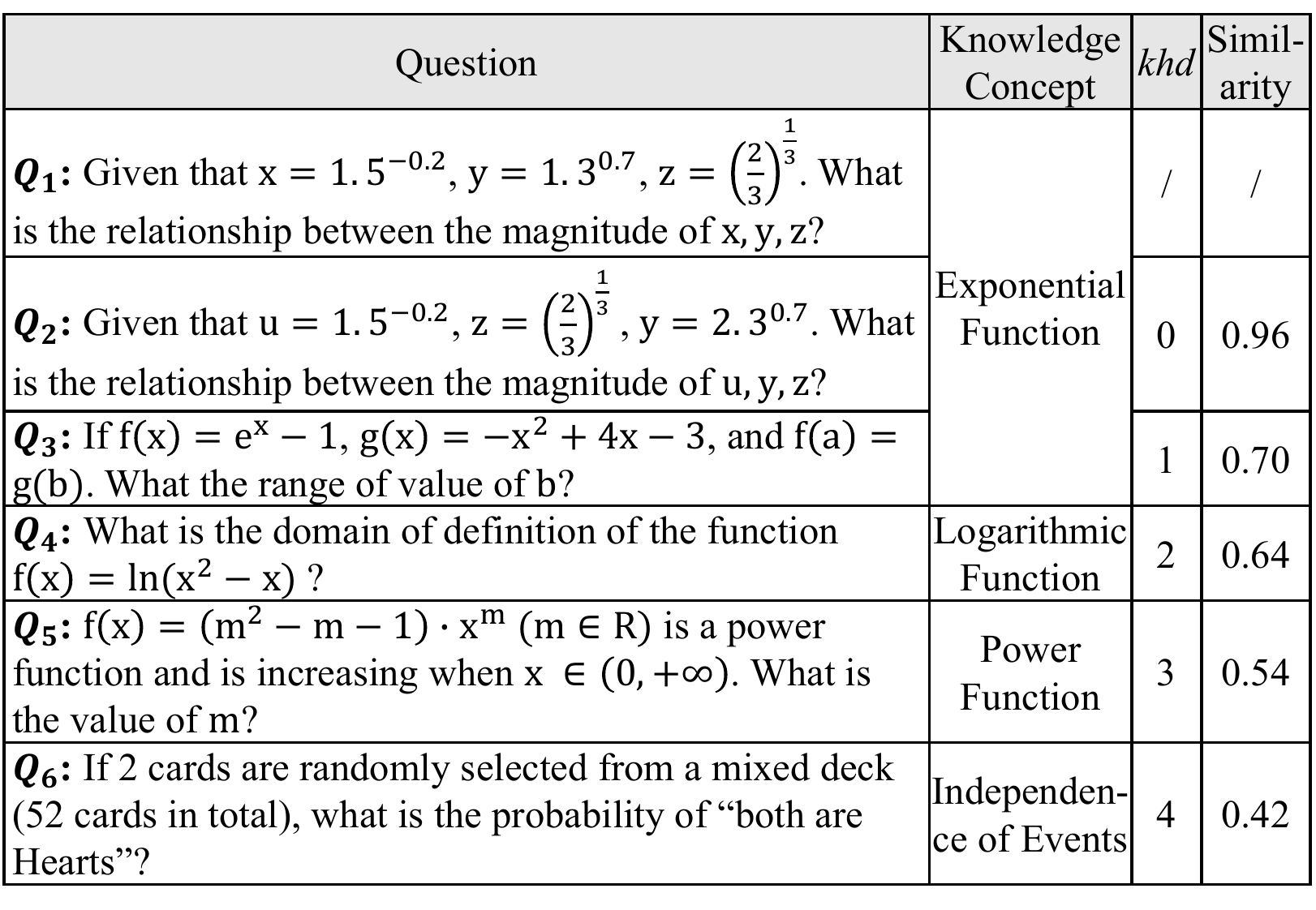} 
\caption{A case of questions with different similarities.}
\label{case}
\end{figure}

Figure~\ref{case} shows a case of questions with different similarities. $Q_1$ is the anchor question. $Q_2$ is the augmented question of $Q_1$ obtained by our question augmentations, which replaces $x$ with $u$, the number $1.3$ with $2.3$ and swaps two clauses. The similarity of $Q_1$ and $Q_2$ is the highest, as their solutions are extremely similar.
$Q_3$ shares the same fine-grained concept with $Q_1$, while solutions are different. Therefore, their similarity is relatively lower.
As the KH-distance $\textit{khd}$ increases, similarities to $Q_1$ decrease. 
$Q_4$ requires different knowledge of \textit{logarithmic functions}, which are the inverses of \textit{exponential functions}. Therefore, $Q_4$ and $Q_1$ are somehow similar. $Q_5$ only shares the same coarse-grained knowledge concept \textit{Function} with $Q_1$, while $Q_6$ is a probability problem without any common knowledge concept with $Q_1$. Thus, they are the most dissimilar. \looseness=-1

\section{Conclusion}

In this paper, we studied the problem of mathematical question understanding. We proposed a novel contrastive pre-training approach, namely QuesCo, to holistically understand mathematical questions.
Specifically, we first designed two-level augmentations for the content and structure of mathematical questions, and generated literally diverse question pairs with mathematically similar purposes. Then, we proposed a novel knowledge hierarchy-aware rank strategy to exploit rich information in knowledge concepts, which generated a fine-grained similarity ranking between questions.
With extensive experiments on three typical downstream tasks in mathematical education, we demonstrated that QuesCo could comprehensively understand mathematical questions, capturing the domain knowledge and holistic purposes of questions. We further conducted more analytical experiments to demonstrate the effectiveness and rationality of QuesCo.
For further study, we hope to generalize our work to more educational questions and exploit more educational properties.
Meanwhile, we will design more intelligent question-based applications for personalized education.

\section{Acknowledgments}
This research was partially supported by grants from the National Natural Science Foundation of China (Grants No. 62106244 and U20A20229), and the Iflytek joint research program.

\bibliography{aaai23}

\end{document}